\documentclass[journal,twoside]{IEEEtran}
\usepackage{multirow}
\usepackage{graphicx}
\usepackage{booktabs}
\usepackage{soul}
\soulregister\cite7
\soulregister\ref7
\usepackage{color}

\usepackage{bbding}

\usepackage{url}

\ifCLASSOPTIONcompsoc
\usepackage[caption=false, font=normalsize, labelfont=sf, textfont=sf]{subfig}
\else
\usepackage[caption=false, font=footnotesize]{subfig}
\fi
\usepackage{amsmath}
\usepackage{amsfonts}

\usepackage{cite}
\usepackage{algorithm}
\usepackage{algpseudocode}

\usepackage{enumerate}
\usepackage{bm}

\usepackage{amsthm}

\begin{document}
\title{Graph Autoencoder Process Monitoring}
\author{Xiangrui~Zhang,~\IEEEmembership{Member,~IEEE}
	\thanks{Xiangrui Zhang is with the School of Information and Control Engineering, China University of Mining and Technology, Xuzhou 221116, China (e-mail: zhangxr@cumt.edu.cn).}
	 % <-this % stops a space
}
\markboth{IEEE Transactions on Neural Networks and Learning Systems}%
{Zhang \MakeLowercase{\textit{et al.}}: Causal Graph Spatial-Temporal Autoencoder for Reliable and Interpretable Process Monitoring}

\maketitle

\begin{abstract}
To improve the reliability and interpretability of industrial process monitoring, this article proposes a Causal Graph Spatial-Temporal Autoencoder (CGSTAE). The network architecture of CGSTAE combines two components: a correlation graph structure learning module based on spatial self-attention mechanism (SSAM) and a spatial-temporal encoder-decoder module utilizing graph convolutional long-short term memory (GCLSTM). The SSAM learns correlation graphs by capturing dynamic relationships between variables, while a novel three-step causal graph structure learning algorithm is introduced to derive a causal graph from these correlation graphs. The algorithm leverages a reverse perspective of causal invariance principle to uncover the invariant causal graph from varying correlations. The spatial-temporal encoder-decoder, built with GCLSTM units, reconstructs time-series process data within a sequence-to-sequence framework. The proposed CGSTAE enables effective process monitoring and fault detection through two statistics in the feature space and residual space. Finally, we validate the effectiveness of CGSTAE in process monitoring through the Tennessee Eastman process and a real-world air separation process.
\end{abstract}


\IEEEpeerreviewmaketitle

\section{Introduction}
\IEEEPARstart{D}{ata-driven} multivariate statistical process monitoring (MSPM) has gained prominence as a powerful tool for monitoring complex industrial processes, ensuring both operation safety and production efficiency\cite{gao2015survey,huang2025EaLDL,DSLAN}. For a given process, MSPM constructs a multivariate statistical model to identify patterns that reflects the system behavior. Once the model is established, control limits are defined based on these patterns under normal operating conditions, which allows for monitoring and detecting deviations from the expected behavior. Over the past decades, MSPM methods based on principal component analysis (PCA)\cite{kano2001new} and canonical variable analysis (CVA)\cite{lu2018sparse} have been extensively studied. With the advent of the artificial intelligence era, deep learning provides more advanced solutions to the MSPM community\cite{kong2021deep}, such as autoencoders (AEs), variational autoencoders (VAEs), long-short term memory networks (LSTMs), and Transformers. Despite these advancements, data-driven MSPM methods are always criticized for their reliability and interpretability\cite{STCG}, leading to a significant gap between laboratory outcomes and industrial applications. For example, changes in operating conditions can degrade the performance of MSPM models, resulting in high false alarm rates. Furthermore, most MSPM models operate as black boxes, making their decision-making processes difficult to explain and leading to skepticism from operators regarding their outputs.

Recently, graph neural networks (GNNs) have received widespread attention for their ability to process graph-structured data using neural networks\cite{wu2020comprehensive}. Commonly used GNNs include graph convolutional network (GCN), graph attention network (GAT), and graph sample and aggregate (GraphSAGE). For a practical industrial process, there often exist physical and chemical connections between variables. By representing these relationships between variables as a graph, GNNs have great potential to enhance the interpretability of MSPM models\cite{jia2025review}. Since the graph structure of a process is typically unknown, graph structure learning is a crucial component of GNN-based MSPM. In existing studies, metric-based approaches, neural approaches, and direct approaches, along with graph regularization techniques on sparsity, smoothness, and community preservation, are widely used for graph structure learning\cite{li2023gslb}. However, the graphs derived from these graph structure learning approaches generally capture correlations rather than causal relationships. As is well-known, correlation does not necessarily imply causation\cite{liu2024information}. Correlation graphs suffer from two inherent drawbacks in process modeling and monitoring. On one hand, correlation graph-based MSPM models are prone to being misled by spurious correlations arising from non-causal factors such as confounders and sample selection bias\cite{zhang2024bayesian}. These spurious correlations may not hold under different operating conditions, thereby compromising the reliability of process monitoring\cite{izmailov2022feature}. On the other hand, correlation graph-based MSPM models are limited to identifying relational associations between variables, which often fail to align with underlying process mechanisms, resulting in poor interpretability. On the contrary, causal graph-based MSPM models are able to provide distinct advantages in terms of both reliability and interpretability\cite{yu2022stable}.

Discovering causal graph consistent with underlying process mechanisms poses a huge challenge. Traditional causal discovery methods can be broadly categorized into constraint-based, score-based, and causal function-based algorithms\cite{cao2022causal}. Constraint-based algorithms, like Peter and Clark (PC) and fast causal inference (FCI), are limited by their inability to distinguish between members of Markov equivalence class. Score-based algorithms, like greedy equivalence search (GES) with Bayesian information criterion score, often overlook the influence of unobservable confounders. Causal function-based algorithms, like linear non-Gaussian acyclic model (LiNGAM) , rely on specific assumptions about the data generation mechanism. Moreover, Granger causality analysis (GCA)\cite{liu2024physics} and transfer entropy (TE)\cite{yu2024intrinsic} are two well-known causal discovery tools for time series, widely used in process modeling and monitoring. However, Granger causality primarily reflects the predictive ability of one variable for another, offering limited insight into the true causal relationships between variables. Similarly, transfer entropy measures information transfer but does not directly capture causal mechanisms. Recent advances in neural networks based causal discovery algorithms, like weight comparison\cite{he2022neural}, continue to prioritize the predictive ability as the foundation for causal discovery, similar to the GCA and TE tools.

To address the aforementioned issues, this article proposes a Causal Graph Spatial-Temporal Autoencoder (CGSTAE) for reliable and interpretable MSPM of industrial processes based on causality. The network architecture of CGSTAE integrates two components: a correlation graph structure learning module based on the spatial self-attention mechanism (SSAM) and a spatial-temporal encoder-decoder module utilizing the graph convolutional long-short term memory (GCLSTM). The SSAM leverages a self-attention mechanism to adaptively learn correlation graphs, thereby capturing dynamic relationships between variables. The causal invariance principle believes that causal relationships remain stable despite changes in correlations\cite{peters2017elements}. Following a reverse perspective of the causal invariance principle, the stable parts of the correlations can be considered as causal relationships. To this end, we design a three-step causal graph structure learning algorithm for causal discovery of industrial processes, which aims to derive a causal graph from the correlation graphs generated by the SSAM. With the derived causal graph, the spatial-temporal encoder-decoder performs reconstruction of time series process data within a sequence-to-sequence framework\cite{Sutskever2014Sequence}, comprising a spatial-temporal encoder and a spatial-temporal decoder, both built with GCLSTM units. Finally, we construct two statistics in both feature space and residual space for process monitoring and fault detection. The main contributions of this article are listed as follows:
\begin{enumerate}[1)]
	\item Model innovation: we propose the CGSTAE model which is composed of a spatial-temporal encoder-decoder module and a spatial-temporal encoder-decoder module to improve the reliability and interpretability of MSPM.
	\item Algorithmic framework: we design a three-step causal graph structure learning algorithm for the CGSTAE model training based on the reverse perspective of the causal invariance principle. The algorithm consists of a pre-training step, a causal graph learning step, and a fine-tuning step.
	\item Causal discovery method: we present a novel causal discovery approach of industrial processes in the causal graph learning step, which discovers the causal graph structure from varying correlations and process knowledge.
\end{enumerate}

\section{Preliminaries}
\subsection{Graph Definition in An Industrial Process}
The interactions between variables in an industrial process can be described by an unweighted directed graph. We define the graph as $\mathcal{G}=\left( \mathcal{V},\mathcal{E} \right)$, where $\mathcal{V}=\left\{ {{v}_{1}},{{v}_{2}},\cdots ,{{v}_{n}} \right\}$ is a set of nodes, and $\mathcal{E}\subseteq \mathcal{V}\times \mathcal{V}$ is a set of directed edges. Typically, the $i$th variable of the given process is considered to be the node ${{v}_{i}}\in \mathcal{V}$ and the associated process data ${{\mathbf{x}}_{i}}$ is regarded as the corresponding node attributes. Moreover, the directed edge $\left( {{v}_{i}},{{v}_{j}} \right)\in \mathcal{E}$ symbolizes the dependency relationship between the variable ${{v}_{i}}$ and the variable ${{v}_{j}}$. Such dependency relationship could be correlation (correlation graph) or causality (causal graph), depending on the graph structure learning approach.
\subsection{GNN-based Process Monitoring}
GNNs have emerged as a powerful tool for process monitoring of complex industrial systems, leveraging their ability to model non-Euclidean data structures and capture intricate relationships between variables\cite{jia2025review}. To represent a given process as a graph, the majority of studies consider each variable as a node for traditional process monitoring\cite{chen2024variational}, while some consider each operation unit or equipment as a node to achieve plant-wide process monitoring\cite{wu2023knowledge}. Furthermore, graph structure learning is the key to GNN-based process monitoring. For metric-based approaches, Chen et al.\cite{chen2021graph} constructed association graph based on the cosine similarity and Euclidean distance between features. Ren et al.\cite{ren2023spatial} utilized mutual information to construct a static graph network snapshot. For neural approaches, Chen et al.\cite{chen2021interaction} transformed the sensor signals into a heterogeneous graph with multiple edge types, and learned the edge weights by the attention mechanism adaptively. For direct approaches, Jia et al.\cite{jia2023graph} attempted to learn the adjacency matrix directly through model training. 

However, all of the above graph structure learning approaches represent the process using a correlation graph rather than a causal graph. Recently, some studies proposed to obtained the causal graph based on process knowledge\cite{he2023causal}. But such process knowledge cannot be guaranteed in terms of completeness and accuracy. There is an urgent need in industrial MSPM for a new causal graph structure learning algorithm.

\section{Causal Graph Spatial-Temporal Autoencoder}
\subsection{Reverse Perspective of Causal Invariance Principle}
The causal invariance principle refers to the idea that causal relationships should remain consistent or invariant under different conditions or transformations\cite{peters2017elements,peters2016causal}. In other words, the underlying causal structure of a system should not change even when the system is viewed or analyzed from different perspectives, or when it is subjected to different interventions or manipulations, as long as those changes do not affect the fundamental causal mechanisms at play. For an industrial process, physical factors related to operating conditions such as temperature, pressure, and raw material quality may fluctuate based on production settings, seasonality, or shifts in machine performance. While these fluctuations can influence observed correlations between variables, causal relationships that are truly fundamental to the system should remain stable across different operation conditions, even if the strength of the correlation changes. Taking a reverse perspective, the stable parts of the correlations can be considered as causal relationships in the case of sufficient fluctuations in data distribution. 

Based on the reverse perspective of the causal invariance principle, we propose the CGSTAE to uncover the invariant causal graph from varying correlations and improve the reliability and interpretability of process monitoring based on causality. Therefore, the causal identifiability conditions we incorporate are: (1) Invariant effect, assuming that the causal effect remains stable across different conditions; and (2) Sufficient intervention, ensuring that the collected data includes adequate interventions.
\subsection{Network Architecture}
\begin{figure}[!htb]
	\centering
	\includegraphics[width=9cm, keepaspectratio]{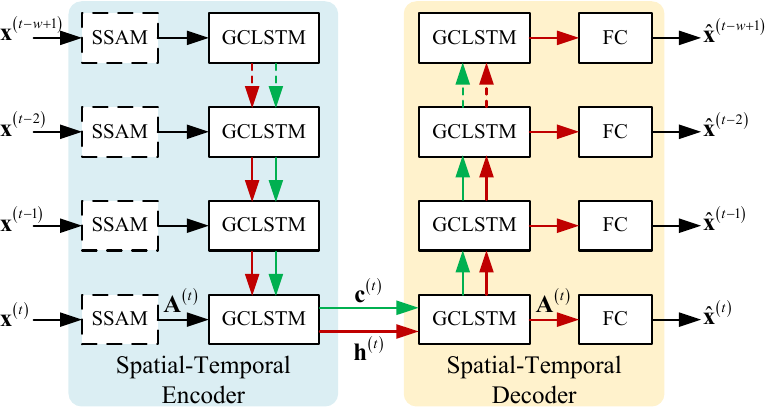}\\
	\caption{Network architecture of CGSTAE.}
	\label{Fig1}
\end{figure}

As shown in Fig. \ref{Fig1}, the architecture of CGSTAE consists of a correlation graph structure learning module based on the SSAM and a spatial-temporal encoder-decoder module based on the GCLSTM. Specifically, the SSAM learns correlation graphs adaptively by the self-attention mechanism to capture the varying correlations between variables. Furthermore, a three-step causal graph structure learning is designed to learn the causal graph from the correlation graphs obtained by SSAM based on the reverse perspective of the causal invariance principle. After learning the causal graph, the correlation graph structure learning module is removed, and the spatial-temporal encoder-decoder module reconstructs the time series process data with GCLSTM by making use of the causal graph.  Finally, we can construct statistics in the feature space and residual space to realize process monitoring and fault diagnosis.
\subsubsection{Correlation Graph Structure Learning Module}
To capture the varying correlations between variables, the SSAM learns correlation graphs adaptively by the self-attention mechanism. Let $\mathbf{X}={{\left[ {{\mathbf{x}}^{\left( 1 \right)}},\cdots ,{{\mathbf{x}}^{\left( N \right)}} \right]}^{T}}$ denote the training data consisting of $N$ normal samples. To account for temporal dynamics, we use sliding windows with length $w$ to reorganize the data. At each time $t$, the input of the model is a data matrix as ${{\mathbf{X}}^{\left( t \right)}}={{\left[ {{\mathbf{x}}^{\left( t-w+1 \right)}},\cdots ,{{\mathbf{x}}^{\left( t \right)}} \right]}^{T}}$. Furthermore, the SSAM takes ${{\mathbf{X}}^{\left( t \right)}}$ as inputs and calculates the similarity via inner product between queries and keys to obtain an attention matrix ${{\mathbf{A}}^{\left( t \right)}}$ as
\begin{equation}
	{{\mathbf{A}}^{\left( t \right)}}=\sigma \left( \frac{{{\left( {{\mathbf{Q}}^{\left( t \right)}} \right)}^{T}}{{\mathbf{K}}^{\left( t \right)}}}{\sqrt{w}} \right)
\end{equation}
where $\sigma $ denotes the sigmoid function and the query matrix ${{\mathbf{Q}}^{t}}$ and the key matrix ${{\mathbf{K}}^{t}}$ are calculated by
\begin{equation}
	\begin{aligned}
		& {{\mathbf{Q}}^{\left( t \right)}}={{\mathbf{X}}^{\left( t \right)}}{{\mathbf{W}}_{\mathbf{Q}}} \\ 
		& {{\mathbf{K}}^{\left( t \right)}}={{\mathbf{X}}^{\left( t \right)}}{{\mathbf{W}}_{\mathbf{K}}} \\ 
	\end{aligned}
\end{equation}
where ${{\mathbf{W}}_{\mathbf{Q}}}$ and ${{\mathbf{W}}_{\mathbf{K}}}$ are two trainable weight matrices of SSAM.

The attention weights calculated by the SSAM represent the variable-to-variable similarities within each sliding window. These similarities can be interpreted as the strengths of correlations among process variables\cite{zhou2024label,chen2022knowledge}. Furthermore, the attention matrix ${{\mathbf{A}}^{\left( t \right)}}$ is regarded as the adjacency matrix of the correlation graph at time $t$. Through the sliding window, SSAM enables adaptive structure learning of correlation graphs.

\subsubsection{Spatial-Temporal Encoder-Decoder Module}
\begin{figure}[!htb]
	\centering
	\includegraphics[width=9cm, keepaspectratio]{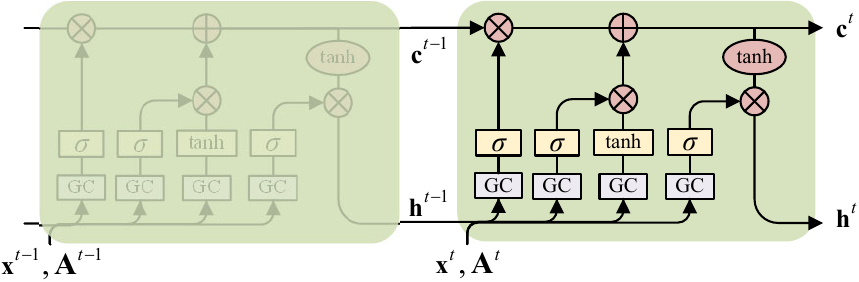}\\
	\caption{GCLSTM unit of the spatial-temporal encoder.}
	\label{Fig2}
\end{figure}

To reconstruct the time series process data, we design a spatial-temporal encoder-decoder module that integrates GCN and LSTM. This module adopts a sequence-to-sequence framework\cite{Sutskever2014Sequence}, comprising a spatial-temporal encoder and a spatial-temporal decoder, both built with GCLSTM units. Following the sequence-to-sequence framework, the spatial-temporal encoder cycles a GCLSTM unit $w$ times in positive order and transmits the final cell state ${{\mathbf{c}}^{\left( t \right)}}$ and hidden state ${{\mathbf{h}}^{\left( t \right)}}$ to the spatial-temporal decoder. Afterwards, spatial-temporal decoder cycles a new GCLSTM unit $w$ times in reverse order, and predicts the reconstruction values with a fully connected (FC) layer. The architecture of the spatial-temporal encoder-decoder module is given in Fig. \ref{Fig1}. In addition, we also depict the GCLSTM unit of the encoder in Fig. \ref{Fig2}.

At each positive step $k\in \left\{ t-w+1,\cdots ,t-1,t \right\}$ within the sliding window, the GCLSTM unit of the encoder updates the cell state and hidden state to extract spatial-temporal features by
\begin{equation}
	\begin{aligned}
		& {{\mathbf{f}}^{\left( k \right)}}=\sigma \left( \text{G}{{\text{C}}_{\mathbf{f}}}\left( \left[ {{\mathbf{x}}^{\left( k \right)}};{{\mathbf{h}}^{\left( k-1 \right)}} \right],{{\mathbf{A}}^{\left( k \right)}} \right)+{{\mathbf{b}}_{\mathbf{f}}} \right) \\ 
		& {{\mathbf{i}}^{\left( k \right)}}=\sigma \left( \text{G}{{\text{C}}_{\mathbf{i}}}\left( \left[ {{\mathbf{x}}^{\left( k \right)}};{{\mathbf{h}}^{\left( k-1 \right)}} \right],{{\mathbf{A}}^{\left( k \right)}} \right)+{{\mathbf{b}}_{\mathbf{i}}} \right) \\ 
		& {{\mathbf{o}}^{\left( k \right)}}=\sigma \left( \text{G}{{\text{C}}_{\mathbf{o}}}\left( \left[ {{\mathbf{x}}^{\left( k \right)}};{{\mathbf{h}}^{\left( k-1 \right)}} \right],{{\mathbf{A}}^{\left( k \right)}} \right)+{{\mathbf{b}}_{\mathbf{o}}} \right) \\ 
		& {{{\mathbf{\tilde{c}}}}^{\left( k \right)}}=\tanh \left( \text{G}{{\text{C}}_{\mathbf{c}}}\left( \left[ {{\mathbf{x}}^{\left( k \right)}};{{\mathbf{h}}^{\left( k-1 \right)}} \right],{{\mathbf{A}}^{\left( k \right)}} \right)+{{\mathbf{b}}_{\mathbf{c}}} \right) \\ 
		& {{\mathbf{c}}^{\left( k \right)}}={{\mathbf{f}}^{\left( k \right)}}\odot {{\mathbf{c}}^{\left( k-1 \right)}}+{{\mathbf{i}}^{\left( k \right)}}\odot {{{\mathbf{\tilde{c}}}}^{\left( k \right)}} \\ 
		& {{\mathbf{h}}^{\left( k \right)}}={{\mathbf{o}}^{\left( k \right)}}\odot \tanh \left( {{\mathbf{c}}^{\left( k \right)}} \right) \\ 
	\end{aligned}
\end{equation}
where ${{\mathbf{f}}^{\left( k \right)}}$, ${{\mathbf{i}}^{\left( k \right)}}$, ${{\mathbf{o}}^{\left( k \right)}}$, ${{\mathbf{c}}^{\left( k \right)}}$, and ${{\mathbf{h}}^{\left( k \right)}}$ represent the forget gate, input gate, output gate, cell state, and hidden state respectively. ${{\mathbf{b}}_{\mathbf{f}}}$, ${{\mathbf{b}}_{\mathbf{i}}}$, ${{\mathbf{b}}_{\mathbf{o}}}$, and ${{\mathbf{b}}_{\mathbf{c}}}$ are the trainable biases. $\text{G}{{\text{C}}_{\mathbf{f}}}$, $\text{G}{{\text{C}}_{\mathbf{i}}}$,$\text{G}{{\text{C}}_{\mathbf{o}}}$, and $\text{G}{{\text{C}}_{\mathbf{c}}}$ denote four graph convolutional (GC) layers, which update the node representations $\mathbf{Z}$ by aggregating neighbor information as follows
\begin{equation}
	\text{GC}\left( \mathbf{Z},{{\mathbf{A}}^{\left( k \right)}} \right)={{\mathbf{D}}^{-\frac{1}{2}}}\left( {{\mathbf{A}}^{\left( k \right)}}+\mathbf{I} \right){{\mathbf{D}}^{-\frac{1}{2}}}\mathbf{Z}{{\mathbf{W}}_{\text{GC}}}
\end{equation}
where $\mathbf{I}$ is the identity matrix with the same dimension as ${{\mathbf{A}}^{\left( k \right)}}$, $\mathbf{D}$ is the degree matrix of $\left( {{\mathbf{A}}^{\left( k \right)}}+\mathbf{I} \right)$, and ${{\mathbf{W}}_{\text{GC}}}$ is a trainable weight matrix.

At each reverse step $k\in \left\{ t,t-1,\cdots ,t-w+1 \right\}$ within the sliding window, the GCLSTM unit of the decoder updates the cell state and hidden state as follows
\begin{equation}
	\begin{aligned}
		& {{\mathbf{f}}^{\left( k-1 \right)}}=\sigma \left( \text{G}{{\text{C}}_{\mathbf{f}}}\left( \left[ {{{\mathbf{\hat{x}}}}^{\left( k \right)}};{{\mathbf{h}}^{\left( k \right)}} \right],{{\mathbf{A}}^{\left( k \right)}} \right)+{{\mathbf{b}}_{\mathbf{f}}} \right) \\ 
		& {{\mathbf{i}}^{\left( k-1 \right)}}=\sigma \left( \text{G}{{\text{C}}_{\mathbf{i}}}\left( \left[ {{{\mathbf{\hat{x}}}}^{\left( k \right)}};{{\mathbf{h}}^{\left( k \right)}} \right],{{\mathbf{A}}^{\left( k \right)}} \right)+{{\mathbf{b}}_{\mathbf{i}}} \right) \\ 
		& {{\mathbf{o}}^{\left( k-1 \right)}}=\sigma \left( \text{G}{{\text{C}}_{\mathbf{o}}}\left( \left[ {{{\mathbf{\hat{x}}}}^{\left( k \right)}};{{\mathbf{h}}^{\left( k \right)}} \right],{{\mathbf{A}}^{\left( k \right)}} \right)+{{\mathbf{b}}_{\mathbf{o}}} \right) \\ 
		& {{{\mathbf{\tilde{c}}}}^{\left( k-1 \right)}}=\tanh \left( \text{G}{{\text{C}}_{\mathbf{c}}}\left( \left[ {{{\mathbf{\hat{x}}}}^{\left( k \right)}};{{\mathbf{h}}^{\left( k \right)}} \right],{{\mathbf{A}}^{\left( k \right)}} \right)+{{\mathbf{b}}_{\mathbf{c}}} \right) \\ 
		& {{\mathbf{c}}^{\left( k-1 \right)}}={{\mathbf{f}}^{\left( k-1 \right)}}\odot {{\mathbf{c}}^{\left( k \right)}}+{{\mathbf{i}}^{\left( k-1 \right)}}\odot {{{\mathbf{\tilde{c}}}}^{\left( k-1 \right)}} \\ 
		& {{\mathbf{h}}^{\left( k-1 \right)}}={{\mathbf{o}}^{\left( k-1 \right)}}\odot \tanh \left( {{\mathbf{c}}^{\left( k-1 \right)}} \right) \\ 
	\end{aligned}
\end{equation}
where the reconstruction value of the process data ${{\mathbf{\hat{x}}}^{\left( k \right)}}$ is predicted by the hidden state and the FC layer as
\begin{equation}
	{{\mathbf{\hat{x}}}^{\left( k \right)}}={{\mathbf{W}}_\text{FC}}{{\mathbf{h}}^{\left( k \right)}}+{{\mathbf{b}}_\text{FC}}
\end{equation}
where ${{\mathbf{W}}_\text{FC}}$ and ${{\mathbf{b}}_\text{FC}}$ are trainable weight and bias of the FC layer.
\subsection{Three-Step Causal Graph Structure Learning}
Taking a reverse perspective of the causal invariance principle, the stable parts of the correlations can be considered as causal relationships in the case of sufficient fluctuations in data distribution. Inspired by this, we propose a three-step causal graph structure learning algorithm for the CGSTAE training to uncover the invariant causal graph from varying correlations. To make it clear, we denote ${{f}_{\text{SSAM}}}$ and ${{f}_{\text{STAE}}}$ as the functions of the correlation graph structure learning module and the spatial-temporal encoder-decoder module, respectively. The trainable model parameters of CGSTAE are represented by $\bm{\theta }=\left[ {{\bm{\theta }}_{\text{SSAM}}},{{\bm{\theta }}_{\text{STAE}}} \right]$, where ${{\bm{\theta }}_{\text{SSAM}}}$ denotes the model parameters of the correlation graph structure learning module and ${{\bm{\theta }}_{\text{STAE}}}$ denotes the model parameters of the spatial-temporal encoder-decoder module.
Algorithm 1 provides detailed pseudocode of the proposed three-step causal graph structure learning.
\begin{algorithm}
	\caption{Three-step causal graph structure learning}\label{alg1}
	\begin{algorithmic}[1]
		\Require Training data $\mathbf{X}$, prior causal graph ${{\mathbf{A}}^{\text{prior}}}$, model settings and optimizer hyperparameters
		\Ensure Model parameters ${{\bm{\theta }}_{\text{STAE}}}$, causal graph $\mathbf{A}$
		\State Pre-training step: initialize $\bm{\theta }=\left[ {{\bm{\theta }}_{\text{SSAM}}},{{\bm{\theta }}_{\text{STAE}}} \right]$
		\For{epoch $=1$ to $ N_{\text{S1}} $}
		\For{batch $=1$ to $ N_{\text{batch}} $}
		\State Forward propagate
		\State Update $\bm{\theta }$ with Eq.(7)
		\EndFor
		\EndFor
		\State Causal graph learning step: freeze ${{\bm{\theta }}_{\text{SSAM}}}$ and ${{\bm{\theta }}_{\text{STAE}}}$
		\For{epoch $=1$ to $ N_{\text{S2}} $}
		\For{batch $=1$ to $ N_{\text{batch}} $}
		\State Forward propagate
		\State Update $\mathbf{A}$ with Eq.(9)
		\EndFor
		\EndFor
		\State Fine-tuning step: remove ${{\bm{\theta }}_{\text{SSAM}}}$ and unfreeze ${{\bm{\theta }}_{\text{STAE}}}$
		\For{epoch $=1$ to $ N_{\text{S3}} $}
		\For{batch $=1$ to $ N_{\text{batch}} $}
		\State Forward propagate
		\State Update ${{\bm{\theta }}_{\text{STAE}}}$ with Eq.(15)
		\EndFor
		\EndFor
	\end{algorithmic}
\end{algorithm}

\subsubsection{Pre-Training Step}
At the first step, the CGSTAE is pre-trained to reconstruct the process data with the mean-squared error (MSE) loss as
\begin{equation}
	\begin{aligned}
		\bm{\theta }&=\arg \underset{\bm{\theta }}{\mathop{\min }}\,{{\mathcal{L}}_{\text{MSE}}}\left( {{\bm{\theta }}_{\text{SSAM}}},{{\bm{\theta }}_{\text{STAE}}} \right) \\ 
%		& =\arg \underset{\bm{\theta }}{\mathop{\min }}\,\frac{1}{\left( N-w+1 \right)w}\sum\limits_{t=w}^{N}{\sum\limits_{k=t-w+1}^{t}{{{\left\| {{{\mathbf{\hat{x}}}}^{\left( k \right)}}-{{\mathbf{x}}^{\left( k \right)}} \right\|}^{2}}}} \\ 
		& =\arg \underset{\bm{\theta }}{\mathop{\min }}\,\sum\limits_{t=w}^{N}{\sum\limits_{k=t-w+1}^{t}{{{\left\| {{{\mathbf{\hat{x}}}}^{\left( k \right)}}-{{\mathbf{x}}^{\left( k \right)}} \right\|}^{2}}}} \\ 
	\end{aligned}
\end{equation}
where the reconstruction values are obtained by both the spatial-temporal encoder-decoder module and correlation graph structure learning module. Therefore, we can represent the reconstruction values of the input data matrix ${{\mathbf{X}}^{\left( t \right)}}$ as
\begin{equation}
	{{\mathbf{\hat{X}}}^{\left( t \right)}}={{f}_{\text{STAE}}}\left( {{\mathbf{X}}^{\left( t \right)}},{{\mathbf{A}}^{\left( t \right)}}={{f}_{\text{SSAM}}}\left( {{\mathbf{X}}^{\left( t \right)}} \right) \right)
\end{equation}
where ${{\mathbf{\hat{X}}}^{\left( t \right)}}={{\left[ {{{\mathbf{\hat{x}}}}^{\left( t-w+1 \right)}},\cdots ,{{{\mathbf{\hat{x}}}}^{\left( t \right)}} \right]}^{T}}$.

After pre-training, the SSAM identifies the varying correlations between variables by learning the correlation graph adaptively.
\subsubsection{Causal Graph Learning Step}
From the correlation graph, the causal graph can be derived using the invariance principle, which asserts that causal relationships remain unchanged even when correlations vary. To this end, we freeze the parameters of both the spatial-temporal encoder-decoder module and correlation graph structure learning module, and introduce a trainable adjacency matrix $\mathbf{A}$ to represent the causal graph. Furthermore, the causal graph can be learned by
\begin{equation}
	\begin{aligned}
		\mathbf{A}& =\arg \underset{\mathbf{A}}{\mathop{\min }}\,\Big({{\mathcal{L}}_{\text{MSE}}}\left( \mathbf{A},{{\bm{\theta }}_{\text{STAE}}} \right)+{{\lambda }_{1}}{{\mathcal{L}}_{\text{invariance}}}\left( \mathbf{A},{{\mathbf{A}}^{\left( t \right)}} \right) \\ 
		& +{{\lambda }_{2}}{{\mathcal{L}}_{\text{prior}}}\left( \mathbf{A},{{\mathbf{A}}^{\text{prior}}} \right)+{{\lambda }_{3}}{{\mathcal{L}}_{\text{sparsity}}}\left( \mathbf{A} \right)+{{\lambda }_{4}}{{\mathcal{L}}_{\text{discrete}}}\left( \mathbf{A} \right) \Big)\\ 
	\end{aligned}
\end{equation}
where ${{\mathcal{L}}_{\text{MSE}}}$ is a MSE term that ensures the reconstruction ability, ${{\mathcal{L}}_{\text{invariance}}}$ is an invariance term for extracting invariant parts of the correlation graph, ${{\mathcal{L}}_{\text{prior}}}$ is a prior term that introduces process knowledge constraints, ${{\mathcal{L}}_{\text{sparsity}}}$ is a sparsity term, and ${{\mathcal{L}}_{\text{discrete}}}$ is a discreteness term. ${{\lambda }_{1}}$, ${{\lambda }_{2}}$, ${{\lambda }_{3}}$, and ${{\lambda }_{4}}$ are four balancing hyperparameters.

For the first term, the causal graph must possess the ability to reconstruct the process data accurately. By freezing the parameters of the entire CGSTAE, the MSE term ensures that the reconstruction capability of the causal graph aligns with that of the correlation graph. The MSE term is expressed as
\begin{equation}
	\begin{aligned}
		{{\mathcal{L}}_{\text{MSE}}}\left( \mathbf{A},{{\bm{\theta 	}}_{\text{STAE}}} \right)=\sum\limits_{t=w}^{N}{\sum\limits_{k=t-w+1}^{t}{{{\left\| {{{\mathbf{\hat{x}}}}^{\left( k \right)}}-{{\mathbf{x}}^{\left( k \right)}} \right\|}^{2}}}}
	\end{aligned}
\end{equation}
where ${{\mathbf{\hat{X}}}^{\left( k \right)}}={{f}_{\text{STAE}}}\left( {{\mathbf{X}}^{\left( t \right)}},\mathbf{A} \right)$.

For the second term, the invariant parts of the correlations are considered causal relationships when data distribution exhibits sufficient fluctuations. To derive the causal graph based on the invariant parts of the correlation graph, the invariance term is introduced. This term penalizes the L1-norm of the adjacency matrix of the residual between causal graph and correlation graph, defined as
\begin{equation}
	{{\mathcal{L}}_{\text{invariance}}}\left( \mathbf{A},{{\mathbf{A}}^{\left( t \right)}} \right)=\sum\limits_{i=1}^{n}{\sum\limits_{j=1}^{n}{\left| {{\mathbf{A}}_{ij}}-\mathbf{A}_{ij}^{t} \right|}}
\end{equation}
where ${{\mathbf{A}}^{t}}={{f}_{\text{SSAM}}}\left( {{\mathbf{X}}^{t}} \right)$.  

For the third term, learning the causal graph from invariant parts of correlation graph requires sufficient distribution changes in the collected data. To overcome the limitations caused by insufficient fluctuations in distribution of the collected data, process knowledge is integrated to constrain the causal graph structure. Such process knowledge refers to the specific insights of human experts on causal relationships among the variables of a specific industrial process. This is achieved through a prior causal graph and its adjacency matrix, denoted as ${{\mathbf{A}}^{\text{prior}}}$, which encapsulates process knowledge about the causal graph. In this matrix,
\begin{itemize}
	\item $\mathbf{A}_{ij}^{\text{prior}}=1$ indicates a causal relationship from variable $i$ to variable $j$,
	\item $\mathbf{A}_{ij}^{\text{prior}}=0$ indicates no causal relationship from variable $i$ to variable $j$, and
	\item $\mathbf{A}_{ij}^{\text{prior}}=\text{NA}$ indicates uncertainty about the causal relationship from variable $i$ to variable $j$.
\end{itemize}

The prior causal graph can be constructed by analyzing process topology, control loops, or leveraging production experience of human experts. To ensure the learned causal graph conforms to the process knowledge, a masked cross-entropy is introduced as a regularization term, defined by
\begin{equation}
	\begin{aligned}
		{{\mathcal{L}}_{\text{prior}}}\left( 	\mathbf{A},{{\mathbf{A}}^{\text{prior}}} \right)&=-\sum\limits_{i=1}^{n}\sum\limits_{j=1}^{n}{{\mathbf{M}}_{ij}}\Big( \mathbf{A}_{ij}^{\text{prior}}\log {{\mathbf{A}}_{ij}}\\
		&+\left( 1-\mathbf{A}_{ij}^{\text{prior}} \right)\log \left( 1-{{\mathbf{A}}_{ij}} \right) \Big)
	\end{aligned}
\end{equation}
where $\mathbf{M}$ is a mask matrix with the same shape as ${{\mathbf{A}}^{\text{prior}}}$, in which ${{\mathbf{M}}_{ij}}=1$ if $\mathbf{A}_{ij}^{\text{prior}}=1/0$, and ${{\mathbf{M}}_{ij}}=0$ if $\mathbf{A}_{ij}^{\text{prior}}=\text{NA}$. This term aligns the learned causal graph with the constraints imposed by ${{\mathbf{A}}^{\text{prior}}}$, ensuring consistency with known causal relationships while allowing flexibility where causal relationships are uncertain.

For the fourth and fifth terms, we introduce constraints to promote sparsity and discreteness of the causal graph. The sparsity is defined by a masked L1-norm of the adjacency matrix as
\begin{equation}
	{{\mathcal{L}}_{\text{sparsity}}}\left( \mathbf{A} \right)=\sum\limits_{i=1}^{n}{\sum\limits_{j=1}^{n}{\left( 1-{{\mathbf{M}}_{ij}} \right){{\mathbf{A}}_{ij}}}}
\end{equation}
The discreteness is defined through an element-wise entropy of the adjacency matrix as
\begin{equation}
	\begin{aligned}
		{{\mathcal{L}}_{\text{discrete}}}\left( \mathbf{A} \right)&=-\sum\limits_{i=1}^{n}\sum\limits_{j=1}^{n}\Big( {{\mathbf{A}}_{ij}}\log {{\mathbf{A}}_{ij}}\\
		&+\left( 1-{{\mathbf{A}}_{ij}} \right)\log \left( 1-{{\mathbf{A}}_{ij}} \right) \Big)
	\end{aligned}
\end{equation}

It is noteworthy that the causal graph represents dynamic causal relationships among process variables in this article, with each node corresponding to a time series observation of a variable. For tractability, the learned causal graph is essentially a temporal stack of causal relationships, capturing all interactions between the process variables across time. Consequently, we do not impose a global acyclic constraint for causal graph learning. For example, loops naturally arise, due to interactions between controlled and manipulated variables under closed-loop control.
\subsubsection{Fine-Tuning Step}
After the causal graph learning step, the adjacency matrix $\mathbf{A}$ of the causal graph is obtained and the correlation graph structure learning module is removed. Furthermore, we freeze the parameters of the correlation graph structure learning module and then fine-tune the parameters of the spatial-temporal encoder-decoder module using learned causal graph, based on the following MSE loss
\begin{equation}
	\begin{aligned}
		{{\bm{\theta }}_{\text{STAE}}}&=\arg \underset{{{\bm{\theta }}_{\text{STAE}}}}{\mathop{\min }}\,{{\mathcal{L}}_{\text{MSE}}}\left( \mathbf{A},{{\bm{\theta }}_{\text{STAE}}} \right) \\ 
		& =\arg \underset{{{\bm{\theta }}_{\text{STAE}}}}{\mathop{\min }}\,\sum\limits_{t=w}^{N}{\sum\limits_{k=t-w+1}^{t}{{{\left\| {{{\mathbf{\hat{x}}}}^{\left( k \right)}}-{{\mathbf{x}}^{\left( k \right)}} \right\|}^{2}}}} \\ 
	\end{aligned}
\end{equation}
where the reconstruction values are obtained by the spatial-temporal encoder-decoder module and the learned causal graph, which can be described by ${{\mathbf{\hat{X}}}^{\left( k \right)}}={{f}_{\text{STAE}}}\left( {{\mathbf{X}}^{\left( t \right)}},\mathbf{A} \right)$.

Finally, we obtain a causal graph-based CGSTAE model for process monitoring, which offers distinct advantages in terms of both reliability and interpretability.
\subsection{CGSTAE-Based Process Monitoring Procedure}
\subsubsection{Fault Detection}
\begin{figure}[!htb]
	\centering
	\includegraphics[width=8cm, keepaspectratio]{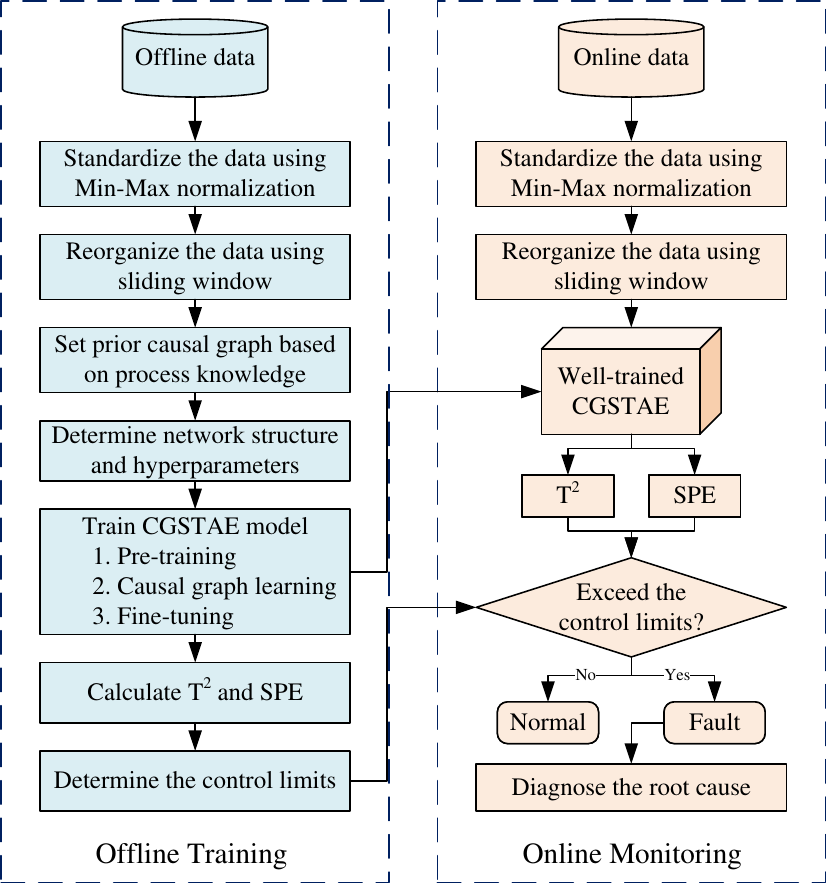}\\
	\caption{Flowchart of the CGSTAE-based process monitoring.}
	\label{Fig3}
\end{figure}

The flowchart of the CGSTAE-based process monitoring procedure is shown in Fig. \ref{Fig3}. During offline modeling phase, the causal graph-based CGSTAE model is trained with $N$ normal training samples. Furthermore, we construct the Hotelling’s t-squared (${{\text{T}}^{2}}$) statistic in the feature space and the squared prediction error (SPE) statistic in the residual space to monitor industrial processes. The final hidden states passed from the encoder to the decoder are used to calculate the ${{\text{T}}^{2}}$ statistic at each time $t$ as
\begin{equation}
	{{\text{T}}^{2}}\left( t \right)={{\left( {{\mathbf{h}}^{\left( t \right)}}-\mathbf{\bar{h}} \right)}^{T}}{{\Sigma }^{-1}}\left( {{\mathbf{h}}^{\left( t \right)}}-{{{\mathbf{\bar{h}}}}^{\left( t \right)}} \right)
\end{equation}
where $\mathbf{\bar{h}}$ and $\Sigma $ denote the mean vector and covariance matrix of the final hidden states.

The reconstruction values within the whole sliding window are used to calculate the SPE statistic at each time $t$ as
\begin{equation}
	\text{SPE}\left( t \right)=\sum\limits_{k=t-w+1}^{t}{{{\left( {{\mathbf{x}}^{\left( k \right)}}-{{{\mathbf{\hat{x}}}}^{\left( k \right)}} \right)}^{T}}\left( {{\mathbf{x}}^{\left( k \right)}}-{{{\mathbf{\hat{x}}}}^{\left( k \right)}} \right)}
\end{equation}

Moreover, the control limits ${{\alpha }_{\text{T2}}}$ and ${{\alpha }_{\text{SPE}}}$ of ${{\text{T}}^{2}}$ and SPE are determined by the kernel density estimation (KDE) with a significance level\cite{zhang2023process}.

During online monitoring phase, at time ${{t}_{\text{new}}}$, the process data is reorganized to get the input matrix ${{\mathbf{X}}^{\left( {{t}_{\text{new}}} \right)}}={{\left[ {{\mathbf{x}}^{\left( {{t}_{\text{new}}}-w+1 \right)}},\cdots ,{{\mathbf{x}}^{\left( {{t}_{\text{new}}} \right)}} \right]}^{T}}$ with the sliding window. Then we can get the two statistics based on the causal graph-based CGSTAE model. Finally, the current process condition is determined by
\begin{equation}
	\left\{ \begin{matrix}
		\begin{aligned}
		&\text{Normal: }{{\text{T}}^{2}}\left( {{t}_{\text{new}}} \right)\le {{\alpha }_{\text{T2}}}\text{ and SPE}\left( {{t}_{\text{new}}} \right)\le {{\alpha }_{\text{SPE}}}  \\
		&\text{Fault: }{{\text{T}}^{2}}\left( {{t}_{\text{new}}} \right)>{{\alpha }_{\text{T2}}}\text{ or SPE}\left( {{t}_{\text{new}}} \right)>{{\alpha }_{\text{SPE}}}  \\
		\end{aligned}
	\end{matrix} \right.
\end{equation}

\subsubsection{Fault Diagnosis}
With the CGSTAE, this article designs an interpretable causal graph-based fault diagnosis method to locate the root cause of a detected fault. Firstly, we calculate the variable contribution by measuring the deviations between measured and reconstructed values of each process variable as
\begin{equation}
	\text{VC}_{i}^{\left( t \right)}=\sum\limits_{k=t-w+1}^{t}{{{\left( x_{i}^{\left( k \right)}-\hat{x}_{i}^{\left( k \right)} \right)}^{2}}}
\end{equation}

By comparing the variable contribution with the SPE threshold ${{\alpha }_{\text{SPE}}}$, we define the variables whose contribution exceeds the threshold as fault variables, and others as normal variables. Afterwards, a set of nodes covering all fault variables is obtained, which is
\begin{equation}
	{{\mathcal{V}}_{\text{fault}}}=\left\{ {{v}_{i}}|\text{VC}_{i}^{\left( t \right)}>{{\alpha }_{\text{SPE}}},t\in \left[ {{t}_{\text{start}}},\cdots ,{{t}_{\text{stop}}} \right] \right\}
\end{equation}

Furthermore, we truncate the adjacency matrix $\mathbf{A}$ of the causal graph using a threshold $\delta $ to obtain a discrete causal graph $\tilde{\mathcal{G}}$ with the adjacency matrix $\mathbf{\tilde{A}}$ as
\begin{equation}
	{{\mathbf{\tilde{A}}}_{ij}}=\left\{ \begin{matrix}
		0,\text{  if }{{\mathbf{A}}_{ij}}\le \delta   \\
		1,\text{  if }{{\mathbf{A}}_{ij}}>\delta   \\
	\end{matrix} \right.
\end{equation}

Finally, we attempt to search for an optimal subgraph on the discrete causal graph $\tilde{\mathcal{G}}$ that includes all fault variables ${{\mathcal{V}}_{\text{fault}}}$ while minimizing the number of normal variables. The optimal subgraph depicts the fault propagation path and consider the source nodes of the subgraph as the potential root causes of the detected fault.
\section{Experiment and Discussion}
To verify the effectiveness of the proposed CGSTAE in process monitoring, we compare it against the following baseline methods using the Tennessee Eastman process and a real-world air separation process:
\begin{itemize}
	\item AE: Consists of an encoder that maps input data into a hidden space and a decoder that reconstructs the input from the hidden representation.
	\item LSTM-AE: Extends the autoencoder by the sequence-to-sequence framework built with LSTM units to process temporal data.
	\item GAE-I: Employs the Pearson correlation coefficient for graph structure learning to uncover correlations between variables, followed by a graph autoencoder for data reconstruction modeling.
	\item GAE-II: Utilizes the transfer entropy for graph structure learning to capture time series information flow between variables, followed by a graph autoencoder for data reconstruction modeling.
	\item DGSTAE: Leverages the SSAM for dynamic graph structure learning and a spatial-temporal graph autoencoder for reconstruction modeling.
	\item KDGCN\cite{guo2023sensor}: Uses process knowledge for graph construction, relationship learning for graph adjustment  and GCN for residual-based fault detection.
	\item KG-GCBiGCN\cite{dong2025knowledge}: Employs expert knowledge for knowledge graph construction, GCN for knowledge-data fusion, and BiGRU for residual-based fault detection.
\end{itemize}

The well-known fault detection rate (FDR) and false alarm rate (FAR) are adopted as performance metrics. Due to the fact that FDR and FAR respectively reflect two independent aspects of the performance, we introduce the F1-score that balances FDR and FAR as the final comprehensive performance metric.
\subsection{Tennessee Eastman Process}
\begin{figure}[!htb]
	\centering
	\includegraphics[width=9cm, keepaspectratio]{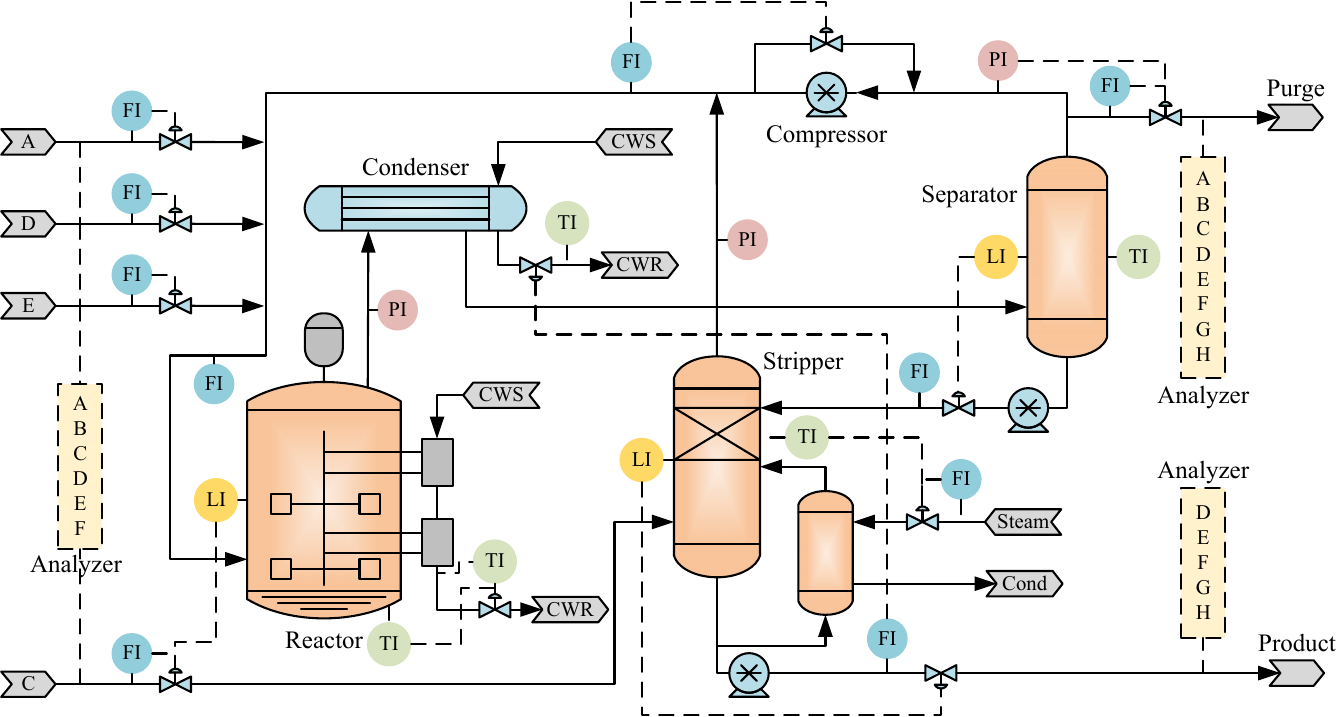}\\
	\caption{Process flow diagram of the TEP.}
	\label{Fig4}
\end{figure}

Tennessee Eastman process (TEP) has been extensively utilized to evaluate the effectiveness of process monitoring techniques\cite{yin2012comparison}. A detailed schematic diagram of the TEP is presented in Fig. \ref{Fig4}, which comprises five operating units: a reactor, a condenser, a compressor, a separator, and a stripper. The TEP involves 41 measured variables, including 22 continuous process measurements and 19 composition measurements, along with 11 manipulated variables. Additionally, the TEP allows for the simulation of 21 faults, facilitating the assessment of monitoring performance. These faults can be categorized as follows: faults 1–7 represent step changes in process variables, faults 8–12 correspond to random variations, fault 13 involves a gradual shift in reaction kinetics, faults 14, 15, and 21 are associated with valve sticking, and faults 16–20 consist of unidentified fault types.

The public dataset of the TEP can be download from the website \footnote{TEP: \url{https://depts.washington.edu/control/LARRY/TE/download.html}}. We select all 52 variables for process monitoring. We use the 960 samples from normal operating conditions as the training set, and 21 fault operating conditions as the testing sets. Each testing set also has 960 samples, and the fault is introduced from the 161st sample. For data reorganization, we set the length of the sliding window $w$ is 5. For CGSTAE model configuration, the dimension of hidden states is set to 2. Four balancing hyperparameters are ${\lambda }_{1}=0.02$, ${\lambda }_{2}=0.08$, ${\lambda }_{3}=0.01$, and ${\lambda }_{4}=0.03$. For CGSTAE model training, the batch size is 32. the learning rate in the pre-training step and fine-tuning step is 0.05, and the learning rate in the causal structure learning step is 0.1. The early stop strategy with a patience of 5 is deployed. For CGSTAE-based process monitoring, the KDE significance level is set to 0.01. The prior causal graph of the TEP is obtained from\cite{wu2022data}.

\begin{table*}[htb]
	\renewcommand{\arraystretch}{1.3}
	\caption{Performance of all methods for the TEP monitoring}
	\label{table_1}
	\centering
	\begin{tabular}{c c c c c c c c c c c c c c c c c}
		\hline
		&\multicolumn{2}{c}{AE}&\multicolumn{2}{c}{LSTM-AE}&\multicolumn{2}{c}{GAE-I}&\multicolumn{2}{c}{GAE-II}&\multicolumn{2}{c}{DGSTAE}&\multicolumn{2}{c}{KDGCN}&\multicolumn{2}{c}{KG-GCBiGCN}&\multicolumn{2}{c}{CGSTAE}\\
		\cmidrule(r){2-3} \cmidrule(r){4-5} \cmidrule(r){6-7} \cmidrule(r){8-9} \cmidrule(r){10-11} \cmidrule(r){12-13}
		\cmidrule(r){14-15}
		\cmidrule(r){16-17}
		Fault&	FDR&	FAR&	FDR&	FAR&	FDR&	FAR&	FDR&	FAR&	FDR&	FAR&	FDR&	FAR&	FDR&	FAR&	FDR&	FAR\\
		\hline
		1&	\textbf{0.998}&	0.063&	0.994&	0.045&	0.993&	0.006&	0.995&	0.013&	0.995&	0.039&	0.983&	0.021&	\textbf{0.998}&	0.019&
		0.996&	0.097\\
		2&	\textbf{0.989}&	0.031&	0.984&	0.039&	0.983&	0.019&	0.988&	0.013&	0.980&	0.000&	0.973&	0.007&	0.988&	0.019&
		0.985&	0.039\\
		3&	0.065&	0.044&	0.055&	0.013&	0.050&	0.038&	0.060&	0.075&	0.031&	0.019&	0.063&	0.079&	0.140&	0.081&
		\textbf{0.179}&	0.129\\
		4&	0.996&	0.038&\textbf{0.999}&	0.039&	0.236&	0.025&	0.469&	0.013&	0.674&	0.006&	0.981&	0.043&	0.944&	0.031&
		\textbf{0.999}&	0.071\\
		5&	0.374&	0.038&	0.374&	0.039&	0.270&	0.025&	0.288&	0.013&	0.275&	0.006&	0.332&	0.043&	\textbf{0.999}&	0.031&
		\textbf{0.999}&	0.071\\
		6&	\textbf{1.000}&	0.019&	0.999&	0.000&	\textbf{1.000}&	0.000&	\textbf{1.000}&	0.013&	0.999&	0.000&	0.988&	0.014&0.993&	0.000&
		0.999&	0.032\\
		7&	\textbf{1.000}&	0.050&	0.999&	0.019&	0.939&	0.006&	\textbf{1.000}&	0.013&	0.999&	0.000&	0.984&	0.000&	\textbf{1.000}&	0.019&
		0.999&	0.019\\
		8&	0.981&	0.050&	0.975&	0.006&	0.973&	0.031&	0.975&	0.025&	0.980&	0.000&	0.963&	0.043&	0.979&	0.050&
		\textbf{0.984}&	0.065\\
		9&	0.079&	0.069&	0.046&	0.052&	0.058&	0.038&	0.059&	0.156&	0.023&	0.058&	0.078&	0.129&	0.106&	0.213&
		\textbf{0.168}&	0.051\\
		10&	0.509&	0.019&	0.504&	0.006&	0.478&	0.000&	0.590&	0.013&	0.743&	0.026&	0.525&	0.000&	0.541&	0.013&
		\textbf{0.880}&	0.019\\
		11&	0.765&	0.081&	0.866&	0.006&	0.476&	0.013&	0.513&	0.019&	0.625&	0.000&	0.874&	0.000&	0.499&	0.019&
		\textbf{0.956}&	0.039\\
		12&	0.990&	0.056&	0.994&	0.026&	0.985&	0.044&	0.985&	0.075&	0.994&	0.006&	0.991&	0.029&	0.986&	0.056&
		\textbf{0.998}&	0.110\\
		13&	0.954&	0.031&	0.954&	0.013&	0.943&	0.000&	\textbf{0.959}&	0.013&	0.949&	0.000&	0.937&	0.000&	0.943&	0.019&
		0.958&	0.045\\
		14&	\textbf{1.000}&	0.063&	0.998&	0.006&	0.999&	0.000&	\textbf{1.000}&	0.013&	0.998&	0.032&	0.988&	0.000&0.995&	0.019&
		0.999&	0.032\\
		15&	0.091&	0.038&	0.076&	0.065&	0.116&	0.006&	0.075&	0.013&	0.053&	0.026&	0.132&	0.014&	0.138&	0.006&
		\textbf{0.219}&	0.045\\
		16&	0.454&	0.056&	0.368&	0.084&	0.270&	0.200&	0.561&	0.231&	0.851&	0.006&	0.342&	0.179&	0.410&	0.244&
		\textbf{0.913}&	0.120\\
		17&	0.945&	0.056&	0.968&	0.019&	0.851&	0.019&	0.881&	0.006&	0.899&	0.000&	0.947&	0.021&	0.814&	0.031&
		\textbf{0.978}&	0.039\\
		18&	0.910&	0.081&	0.904&	0.071&	0.896&	0.013&	0.899&	0.019&	0.900&	0.000&	0.896&	0.000&	0.916&	0.025&
		\textbf{0.919}&	0.090\\
		19&	0.230&	0.031&	0.503&	0.013&	0.126&	0.013&	0.149&	0.013&	0.583&	0.000&	0.128&	0.000&	0.209&	0.019&
		\textbf{0.679}&	0.065\\
		20&	0.565&	0.044&	0.633&	0.000&	0.493&	0.013&	0.671&	0.000&	0.775&	0.000&	0.570&	0.000&	0.473&	0.019&
		\textbf{0.814}&	0.000\\
		21&	0.506&	0.056&	0.493&	0.039&	0.416&	0.031&	0.461&	0.038&	0.375&	0.026&	0.431&	0.071&	0.470&	0.019&
		\textbf{0.645}&	0.068\\
		\hline
		Average	&0.686&	0.048&	0.699&	0.029&	0.598&	0.026&	0.646&	0.037&	0.700&	\textbf{0.012}&0.672&	0.033&	0.692&	0.045&
		\textbf{0.822}&	0.059\\
		F1-score&\multicolumn{2}{c}{0.809}&\multicolumn{2}{c}{0.820}&\multicolumn{2}{c}{0.746}&\multicolumn{2}{c}{0.782}&\multicolumn{2}{c}{0.822}&\multicolumn{2}{c}{0.801}&\multicolumn{2}{c}{0.814}&\multicolumn{2}{c}{\textbf{0.896}}\\
		\hline
	\end{tabular}
\end{table*}

The performance of all methods for the TEP monitoring is listed in Table \ref{table_1}, evaluated by FDR, FAR, and F1-score. Among all the methods, CGSTAE achieves the best overall performance with the highest F1-score of 0.896, effectively maximizing detection accuracy while maintaining low false positives. Leveraging causal graph learning and spatial-temporal modeling, CGSTAE emerges as the most reliable method for TEP monitoring, achieving superior detection accuracy, particularly for challenging faults (e.g., faults 9-11, 15-17, 19-21). DGSTAE also performs well, with an F1-score of 0.822 and the lowest FAR, underscoring the importance of spatial-temporal graph learning. KDGCN and KG-GCBiGCN deliver satisfactory results, with F1-scores of 0.801 and 0.814, respectively, demonstrating the effectiveness of integrating process knowledge. Furthermore, LSTM-AE achieves an F1-score of 0.820, indicating its ability to handle temporal dependencies effectively. In contrast, GAE-I, GAE-II, and AE exhibit subpar monitoring performance.

\begin{figure}[!htb]
	\centering
	\subfloat[]{\includegraphics[width=4.5cm]{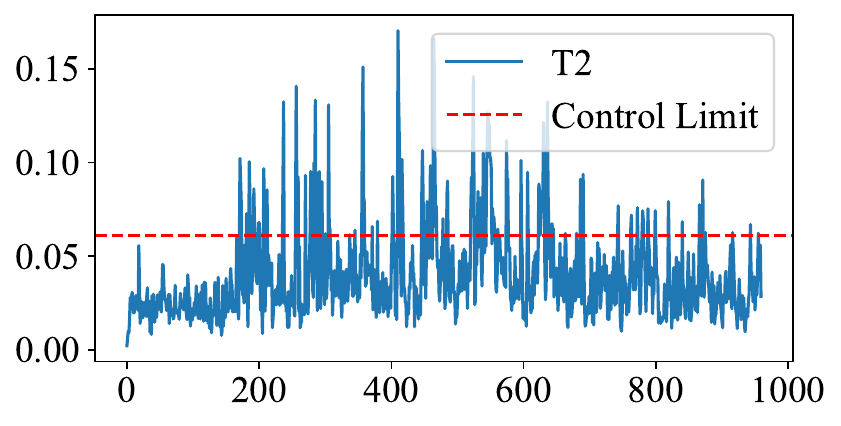}}
	\subfloat[]{\includegraphics[width=4.5cm]{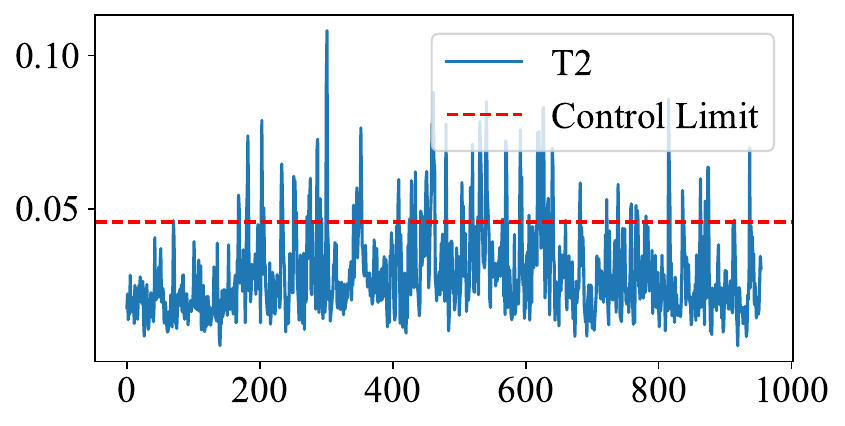}}
	
	\subfloat[]{\includegraphics[width=4.5cm]{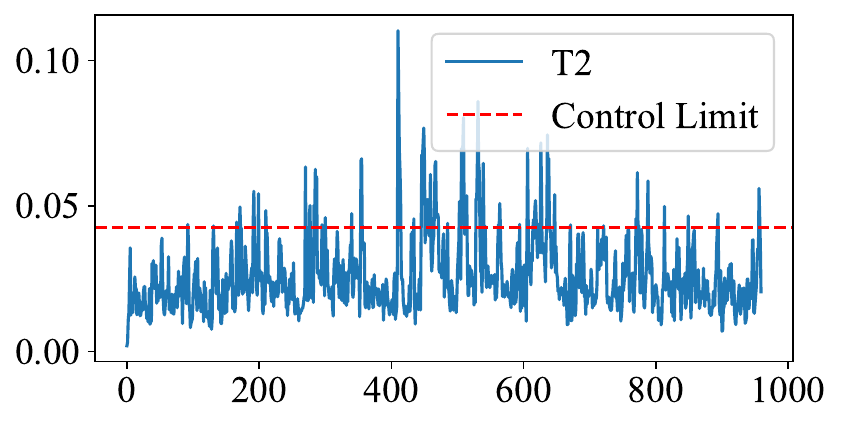}}
	\subfloat[]{\includegraphics[width=4.5cm]{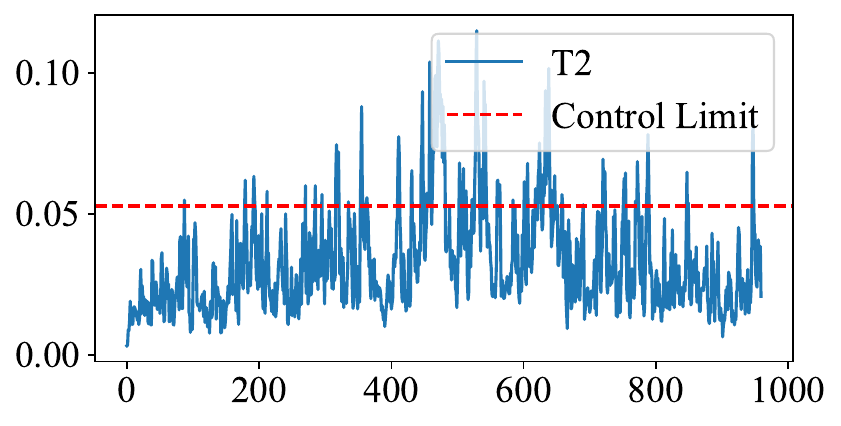}}
	
	\subfloat[]{\includegraphics[width=4.5cm]{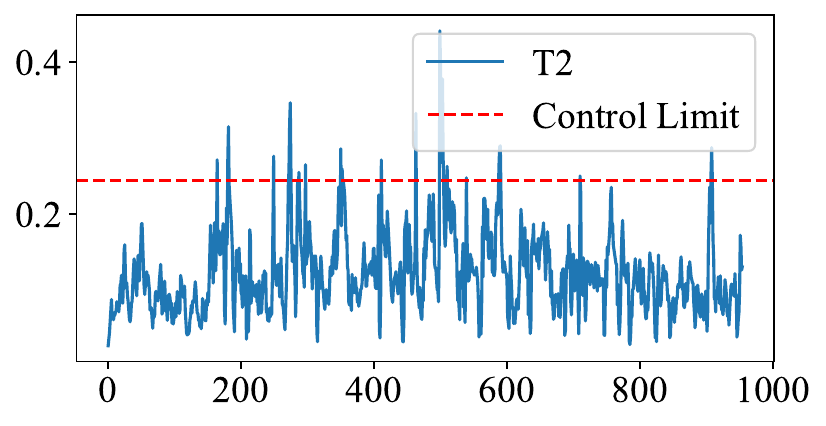}}		
	\subfloat[]{\includegraphics[width=4.5cm]{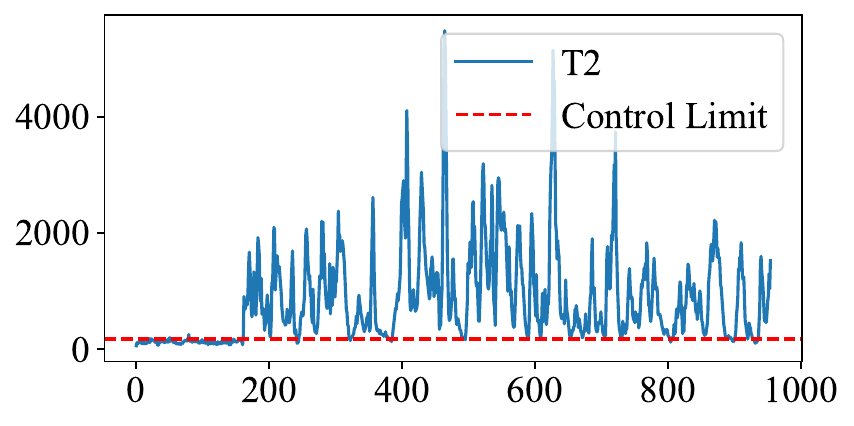}}
	\caption{${{\text{T}}^{2}}$ statistic of different methods for fault 11 of TEP. (a) AE (b) LSTM-AE (c) GAE-I (d) GAE-II (e) DGSTAE (f) CGSTAE. The residual-based fault detection methods KDGCN and KG-GCBiGCN do not have ${{\text{T}}^{2}}$ statistic.}
	\label{Fig5}
\end{figure}
\begin{figure}[!htb]
	\centering
	\subfloat[]{\includegraphics[width=4.5cm]{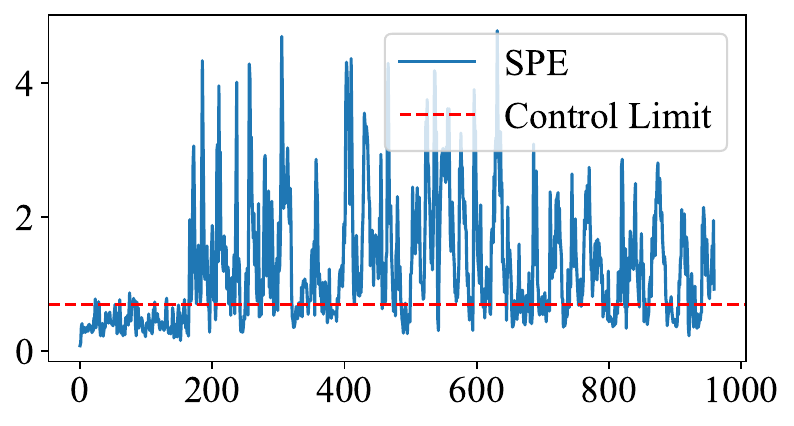}}
	\subfloat[]{\includegraphics[width=4.5cm]{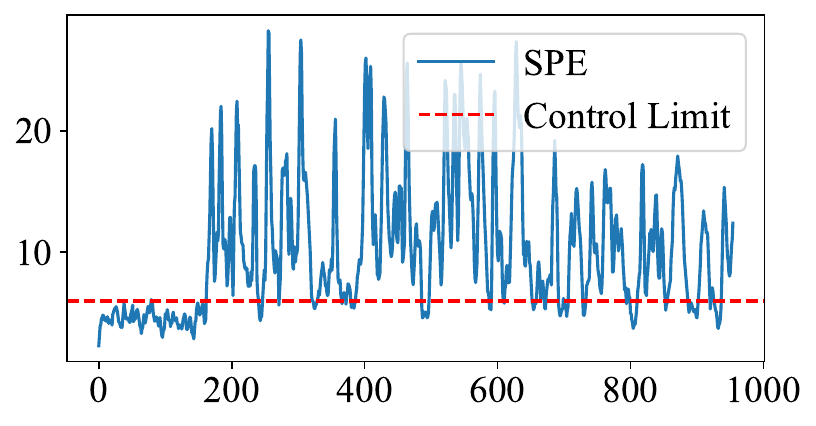}}
	
	\subfloat[]{\includegraphics[width=4.5cm]{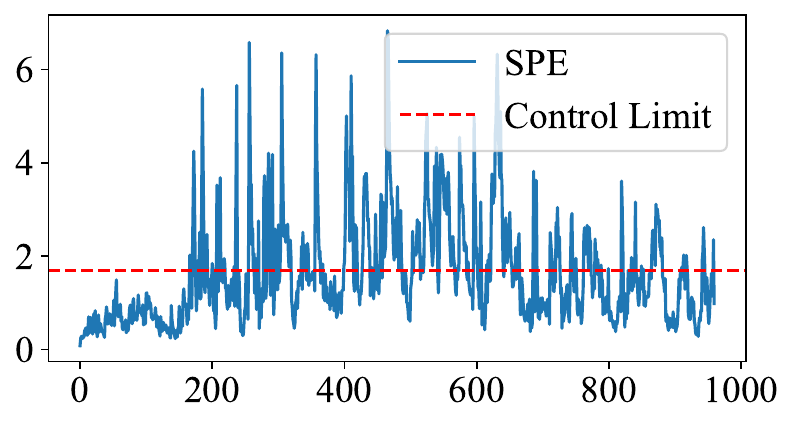}}
	\subfloat[]{\includegraphics[width=4.5cm]{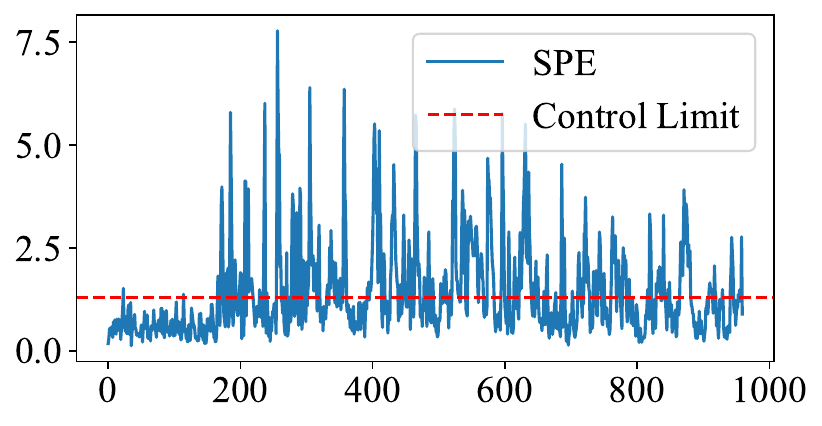}}
	
	\subfloat[]{\includegraphics[width=4.5cm]{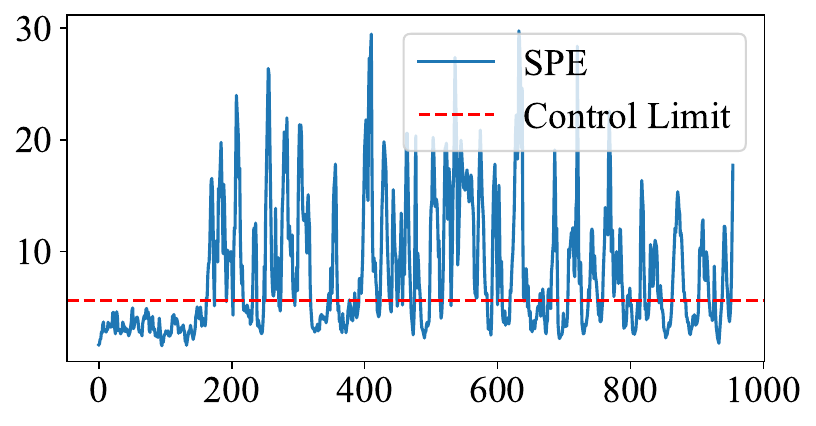}}
	\subfloat[]{\includegraphics[width=4.5cm]{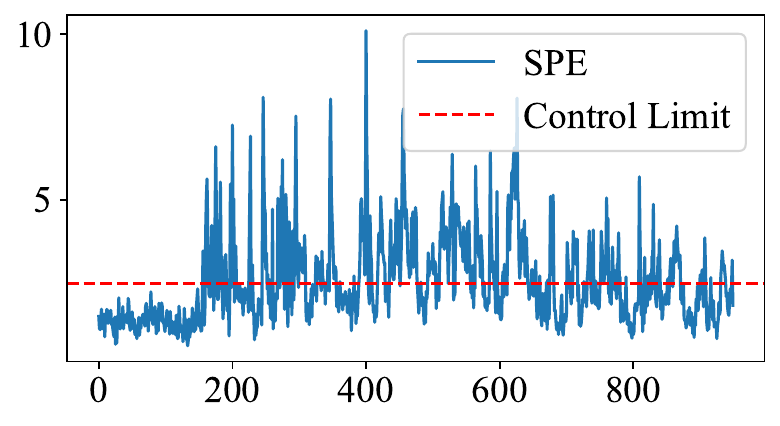}}
	
	\subfloat[]{\includegraphics[width=4.5cm]{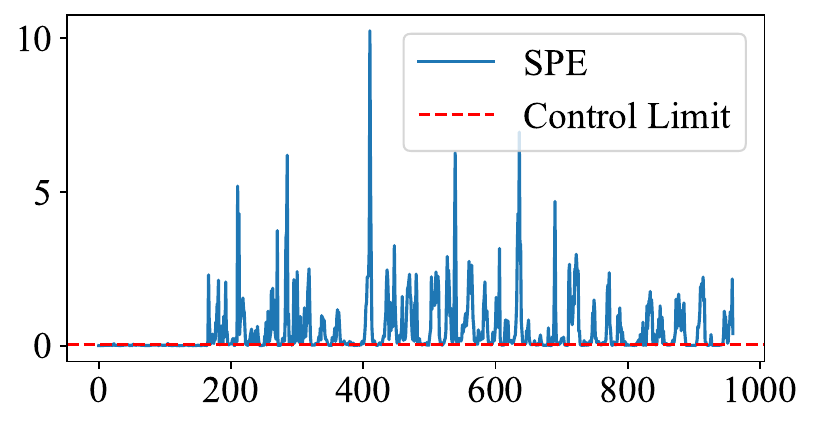}}
	\subfloat[]{\includegraphics[width=4.5cm]{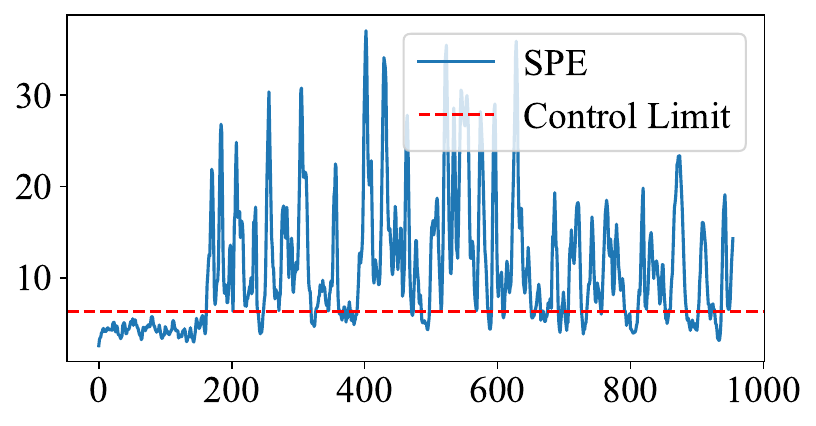}}
	\caption{SPE statistic of different methods for fault 11 of TEP. (a) AE (b) LSTM-AE (c) GAE-I (d) GAE-II (e) DGSTAE (f) KDGCN (g) KG-GCBiGCN (h) CGSTAE.}
	\label{Fig6}
\end{figure}

\begin{figure}[!htb]
	\centering
	\subfloat[]{\includegraphics[width=4.5cm]{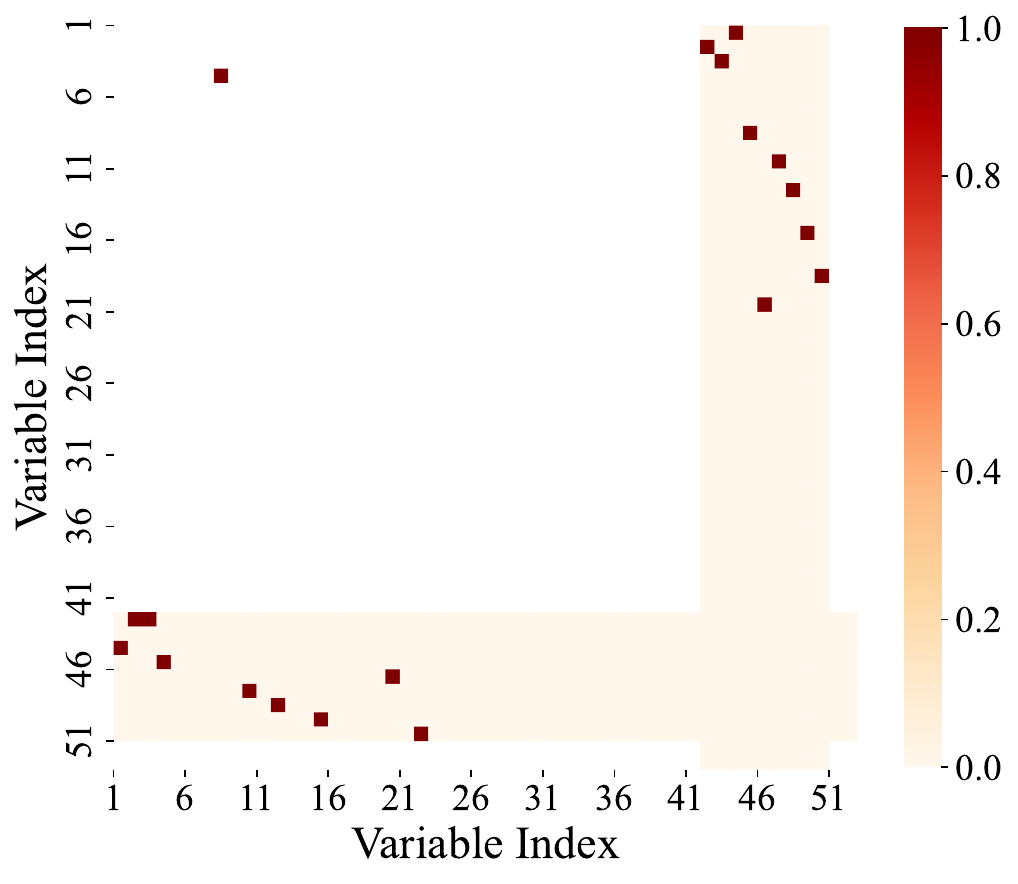}}
	\subfloat[]{\includegraphics[width=4.5cm]{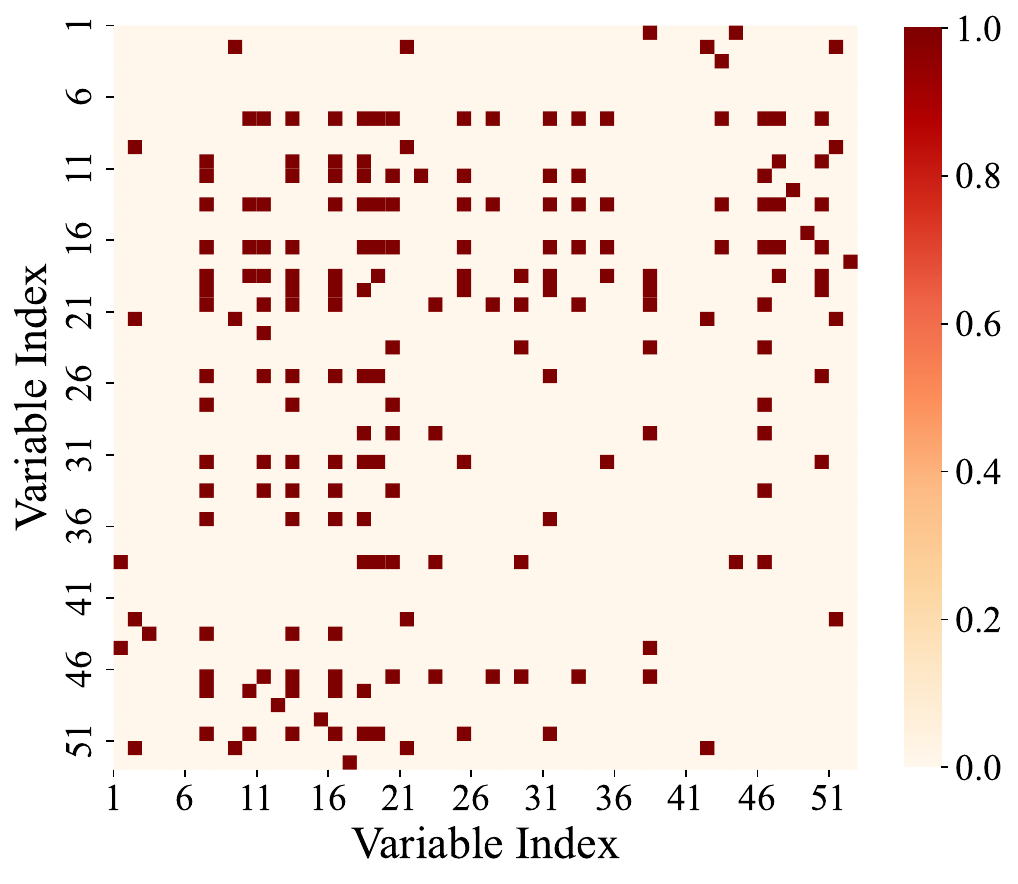}}
	
	\subfloat[]{\includegraphics[width=4.5cm]{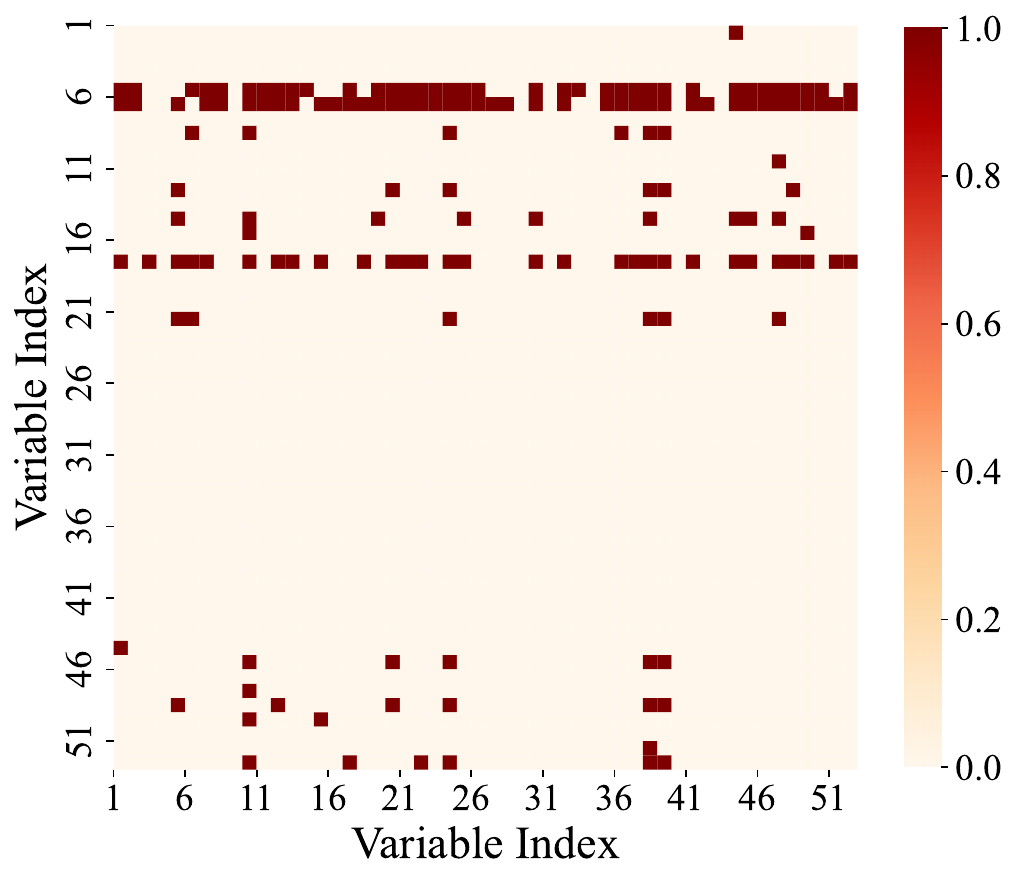}}
	\subfloat[]{\includegraphics[width=4.5cm]{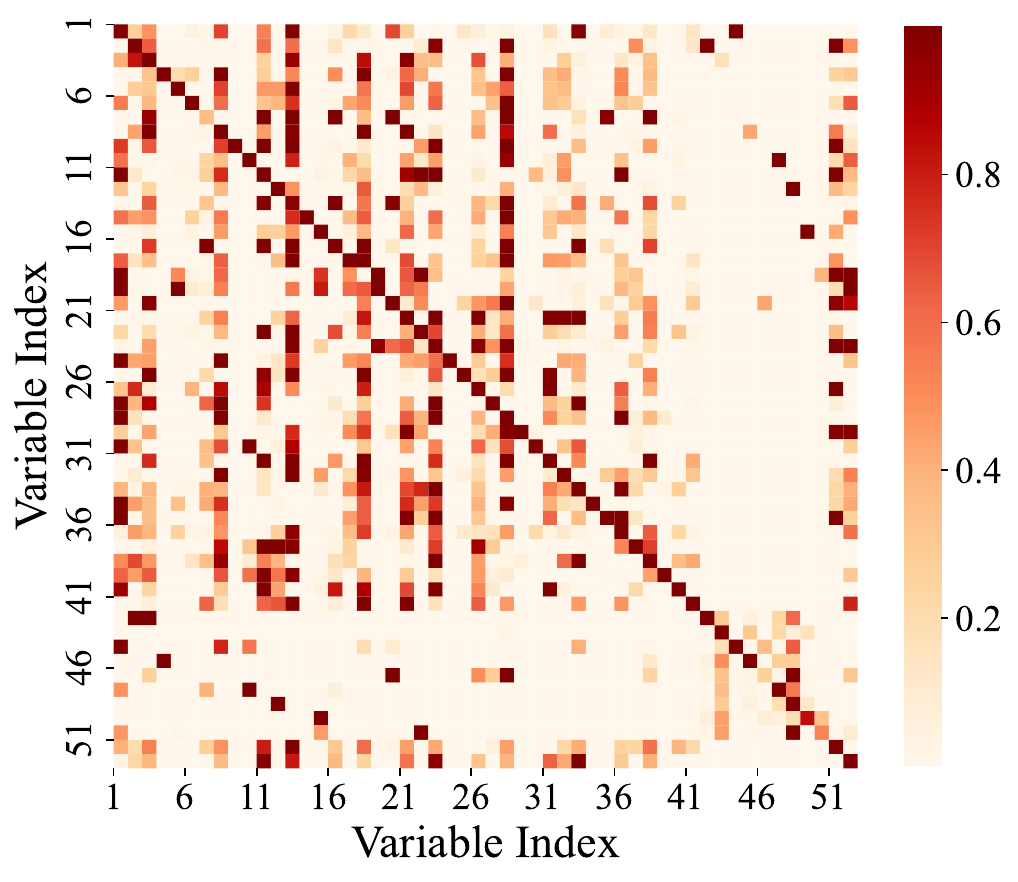}}
	\caption{Graph structures of the TEP. (a) Prior causal graph (b) GAE-I (c) GAE-II (d) CGSTAE.}
	\label{Fig7}
\end{figure}

Taking fault 11 in as an example, we present the ${\text{T}}^{2}$ and SPE statistics of different methods on the testing data in Figs. \ref{Fig5} and \ref{Fig6} to compare the monitoring performance. The red dashed lines represent the control limits, and the fault is introduced from the 161st sample. The ${\text{T}}^{2}$ statistics of baseline methods, including AE, LSTM-AE, GAE-I, GAE-II, and DGSTAE, are difficult to distinguish between faulty and normal operating conditions. As a comparison, the statistics of the proposed CGSTAE can clearly distinguish them. Moreover, the SPE statistics of these baseline methods are below the control limits at many points after the fault occurs, resulting in a lower detection rate. The proposed CGSTAE successfully detects the fault 11 of the TEP by both ${\text{T}}^{2}$ and SPE statistics while maintaining a lower rate of false alarms. 

Fig. \ref{Fig7}(a) shows the prior causal graph of the TEP derived from process knowledge. Figs. \ref{Fig7}(b-d) display the graph structure learning results of GAE-I, GAE-II, and CGSTAE. It is evident that both the Pearson correlation coefficient and transfer entropy struggle to identify the true causal relationships that align with the prior causal graph from process knowledge. Thanks to the causal graph structure learning algorithm, the causal graph learned by CGSTAE is consistent with underlying process mechanisms, greatly improving the interpretability of the model.

\section{Conclusion}
This article proposes the CGSTAE for reliable and interpretable process monitoring, which combines a correlation graph structure learning module based on SSAM and a spatial-temporal encoder-decoder module utilizing GCLSTM. Leveraging a reverse perspective of the causal invariance principle, we present a novel three-step causal graph structure learning algorithm. The effectiveness of CGSTAE in process monitoring is validated through the Tennessee Eastman process and a real-world air separation process. In the future, we can further consider practical issues such as multiple sampling rates\cite{song2025soft,chang2022flexible} and process drift in industrial processes.

\ifCLASSOPTIONcaptionsoff
\newpage
\fi

\bibliographystyle{IEEEtran}
\bibliography{IEEEabrv,reference}

\end{document}